\documentclass[twoside]{article}
\usepackage[accepted]{aistats2014}
\usepackage[parfill]{parskip}  
\usepackage{graphicx}
\usepackage{algorithm}
\usepackage{algorithmic}

\usepackage{amssymb}
\usepackage{caption}
\usepackage{subcaption}
\usepackage{epstopdf}
\usepackage{listings}
\usepackage{color}
\usepackage{balance}

\lstnewenvironment{bodycode}[2]{\small\lstset{caption=#1,label=#2,basicstyle=\small\ttfamily}}{}
\lstnewenvironment{code}[2]{\small\lstset{caption=#1,label=#2}}{}

\newenvironment{inline}[1]{{\small\ttfamily #1}}{}

\usepackage[numbers,square]{natbib}
\usepackage[linktocpage=true]{hyperref}

\renewcommand\vdots{%
  \vbox{\baselineskip2pt\lineskiplimit0pt\kern1pt\hbox{.}\hbox{.}\hbox{.}\kern-1pt}}

\def \commit {1fc10eab278d6a3795bd230d40408f70e88fccaf}
\def \ucommit {8b5734c778b748fc3baa61c5f10bb3158b838582}
\def \gcommit {1fc10eab278d6a3795bd230d40408f70e88fccaf}

\begin{document}

\twocolumn[

%\aistatstitle{Incremental Interpretation of Probabilistic Programs}

\aistatstitle{A New Approach to Probabilistic Programming Inference}

%====
%
%PPGPP: Parallel Particle Gibbs for Probabilistic Programs
%
%Fast, Parallel Inference for Probabilistic Programs via Particle Gibbs
%
%====

%\aistatsauthor{ Anonymous Author 1 \And Anonymous Author 2 \And Anonymous Author 3 }

%\aistatsaddress{ Unknown Institution 1 \And Unknown Institution 2 \And Unknown Institution 3 }

\aistatsauthor{ Frank Wood \And Jan Willem van de Meent \And Vikash Mansinghka }

\aistatsaddress{ Department of Engineering \\University of Oxford \And  Department of Statistics \\ Columbia University \And Computer Science \& AI Lab \\ Massachusetts Institute of Technology } 
]

%\author{
%Frank Wood\thanks{\url{http://www.robots.oc.ac.uk/~fwood}} \\
%Department of Engineering Science\\
%University of Oxford\\
%Oxford, UK OX1 3PJ \\
%\texttt{fwood@robots.ox.ac.uk} \\
%\And
%Vikash Mansinghka \\
%Department of Brain and Cognitive Sciences\\
%Massachusetts Institute of Technology \\
%Cambridge, MA 02139 \\
%\texttt{vkm@mit.edu} \\
%
%}

% The \author macro works with any number of authors. There are two commands
% used to separate the names and addresses of multiple authors: \And and \AND.
%
% Using \And between authors leaves it to \LaTeX{} to determine where to break
% the lines. Using \AND forces a linebreak at that point. So, if \LaTeX{}
% puts 3 of 4 authors names on the first line, and the last on the second
% line, try using \AND instead of \And before the third author name.

\begin{abstract}
%The abstract paragraph should be indented 1/2~inch (3~picas) on both left and
%right-hand margins. Use 10~point type, with a vertical spacing of 11~points.
%The word \textbf{Abstract} must be centered, bold, and in point size 12. Two
%line spaces precede the abstract. The abstract must be limited to one
%paragraph.
% !TEX root =  main.tex

We introduce and demonstrate a new approach to inference in expressive
probabilistic programming languages based on particle Markov chain
Monte Carlo. Our approach is simple to implement and easy to parallelize.
It applies to Turing-complete probabilistic programming languages and
supports accurate inference in models that make use of complex control
flow, including stochastic recursion.  It also includes primitives from
Bayesian nonparametric statistics. Our experiments show that this
approach can be more efficient than previously introduced single-site
Metropolis-Hastings methods.

%We introduce and demonstrate a new approach to probabilistic programming inference based on particle Markov chain Monte Carlo that is more efficient than single-site Metropolis Hastings for at least some programs that correspond to expressive models with rich conditional dependency structure.   This approach is intuitive, easy to implement, and nearly trivially parallelizable.

\end{abstract}

\section{Introduction}
% !TEX root =  main.tex

%Generative models specified via data-generating programs are quite powerful in  they are extremely compact yet are capable of encoding models of extremely high expressivity.  Such probabilistic programs are compact in two senses: short programs are capable of describing state-of-the-art probabilistic models, and, via mechanisms including recursion, programs can efficiently describe exceedingly expressive state machine architectures.

%Probabilistic programming languages enable the description of
%probabilistic models via executable code. Each latent variable in a
%probabilistic model is represented by the invocation of a random
%(sampling) procedure. The data to be explained is represented by
%constraints on the program that force the value of some of the latent
%variables. Inference engines for probabilistic programming languages
%attempt to explore or otherwise characterize the distribution on
%execution traces of this code, conditioned on the forced latent
%variables. One common approach to inference is to sample execution
%traces from some distribution that approximates the conditional
%distribution; doing this effectively, for arbitrary probabilistic
%programs written in an expressive language, is a significant
%challenge.

Probabilistic programming differs substantially from traditional
programming.  In particular, probabilistic programs are written with 
parts not fixed in advance that instead take values generated at runtime
by random sampling procedures.  Inference in probabilistic
programming characterizes the conditional distribution of such 
variables given observed data assumed to have been generated by executing the probabilistic program.  Exploring the joint distribution over program execution traces that could have generated the observed
 data using Markov 
chain sampling techniques is one way to produce such a characterization. 

%essence of probabilistic programming inference.%; otherwise known in traditional probabilistic modelling as observed data.

We propose a novel combination of ideas from probabilistic
programming \cite{goodman2012church} and particle Markov chain Monte
Carlo (PMCMC) \cite{andrieu2010particle} that yields a new
scheme for exploring and characterizing the set of probable execution traces.
%set of plausible execution traces. 
As our approach is based on repeated simulation of the probabilistic
program, it is easy to implement and parallelize. We show that our
approach supports accurate inference in models that make use of
complex control flow, including stochastic recursion, as well as
primitives from nonparametric Bayesian statistics. Our experiments
also show that this approach can be more efficient than previously
introduced single-site Metropolis-Hastings (MH) samplers \cite{wingate2011lightweight}.

\section{Language}
% !TEX root =  main.tex

%\subsection{Syntax}
The probabilistic programming system Anglican\footnote{\url{http://www.robots.ox.ac.uk/~fwood/anglican/}} exists in two versions. The results in this paper were obtained using what is now called {\em interpreted} Anglican\footnote{\url{http://bitbucket.org/probprog/interpreted-anglican}} which employed an interpreted execution model and a language syntax derived from the Venture\footnote{\url{http://probcomp.csail.mit.edu/venture/}} modeling language. Since the time of original publication, both the syntax and execution model have been changed.  The Venture style syntax is now deprecated, with the new version using a syntax much closer to that of Anglican's host language Clojure. This new Anglican version\footnote{\url{http://bitbucket.org/probprog/anglican}}, simply called Anglican, is a compiled language;  a continuation-passing style (CPS) transformation compiles Anglican programs into Clojure programs that are then subsequently compiled to Java Virtual Machine (JVM) bytecode by the Clojure just-in-time compiler. 

The change in language execution model and syntax affect neither the substance or the claims of this paper, however, readers wishing to experiment with the language starting from the code examples appearing in Section~\ref{sec:tests} would do well to note this evolution.  What does change is the absolute time required to perform inference in the given models.  In general, the latest compiled version of Anglican is ten to one hundred times faster than its interpreted ancestor.

\subsection{Original Venture-Style Syntax}

The original, deprecated Anglican syntax is a Scheme/Lisp-dialect that is extended with three top-level special forms that we refer to as directives

\begin{bodycode}{}{}
[assume symbol <expr>]
[observe <expr> <const>]
[predict <expr>]
\end{bodycode}
Here, each \inline{<expr>} is a Scheme/Lisp-syntax expression, \inline{symbol} is a unique symbol, and \inline{<const>} is a constant-valued deterministic expression.   
%Note that the phrases ``evaluation of an expression'' and ``application of a function to its arguments'' should be understood as Scheme/Lisp terms of art \cite{sicp}.  
%For readers not familiar with functional programming common expression syntax \inline{(<proc> <arg> $\ldots$ <arg>)} where each of \inline{<proc>} and all \inline{<arg>}'s may be expressions themselves with the restriction that \inline{<proc>} must evaluate to a procedure that can be applied to the given set of arguments.

%Other supported expression syntaxes include the   %, particularly in Sec.~\ref{sec:experiments}.

Semantically \inline{assume}'s are (random) variable (generative) declarations, \inline{observe}'s condition the distribution of \inline{assume}'d variables by (noisily) constraining the output values of (random) functions of \inline{assume}'d variables to match observed data, and \inline{predict}'s are ``watches'' which report on (via print out) the values of variables in program traces as they are explored.  In Anglican, probabilistic program interpretation is taken to be a forever-continuing exploration of the space of execution traces that obey (where hard) or reflect (where soft) the \inline{observe}'d constraints in order to report functions (via \inline{predict}'s) of the conditional distribution of subsets of the \inline{assume}'d variables.  

Like Scheme/Lisp, Anglican eagerly and exchangeably recursively ``evaluates'' subexpressions, for instance of \inline{<expr> = (<proc> <arg> $\ldots$ <arg>)}, before ``applying'' the procedure (which may be a random procedure or special form) resulting from evaluating the first expression \inline{<proc>}  to the value of its arguments.   Anglican counts applications, reported as a  computational cost proxy in Sec~\ref{sec:tests}.    Anglican supports several special forms, notably  \inline{(lambda (<arg> $\ldots$ <arg>) <body>))} which allows creation of new procedures and \inline{(if <pred> <cons> <alt>)} which supplies branching control flow; also \inline{begin}, \inline{let}, \inline{define}, \inline{quote}, and \inline{cond}.  Anglican exposes \inline{eval} and \inline{apply}.  Built-in  deterministic procedures include \inline{list}, \inline{car}, \inline{cdr}, \inline{cons},  \inline{mem}, etc., and arithmetic procedures \inline{+}, \inline{-}, \inline{$\backslash$},  etc.   

All randomness in the language originates from built-in ``random primitives'' of which there are two types. Elementary random primitives, such as \inline{poisson}, \inline{gamma}, \inline{flip}, \inline{discrete}, \inline{categorical}, and \inline{normal}, generate independent and identically distributed (i.i.d.)~samples when called repeatedly with the same arguments. Exchangeable random primitives, such as \inline{crp} or \inline{beta-bernoulli}, return a random procedure with internal state that generates an exchangeably distributed samples when called repeatedly.

 In the interpreted version of Anglican the outer \inline{<proc>} in all \inline{observe} \inline{<expr>}'s must be a built-in random primitive, which guarantees that likelihood of output given arguments can be computed exactly. 

\subsection{New Clojure-Style Syntax}

The compiled version of Anglican now supports a syntax that more closely integrates with the host language Clojure. The macro {\tt defquery} defines an Anglican program from within Clojure

\begin{bodycode}{}{}
(defquery symbol [arg1 arg2 ...] <body>)
\end{bodycode}
Here \inline{symbol} is the name of the program, and the arguments may be used to pass values of parameters or observed variables to the program. The set of allowable body expressions \inline{<body>} is a subset of the Clojure language, in which all basic Clojure language forms and first order primitives are supported. Macros and higher-order functions are not inherited from Clojure, although a subset of macros\footnote{\inline{when, cond}} and higher-order functions\footnote{\inline{map, reduce, filter, some, repeatedly, comp, partial}} has been implemented. 

In the CPS version of Anglican, random primitives such as \inline{normal} and \inline{discrete} return first class distribution objects, rather than sampled values, and, furthermore, are user programmable without requiring modifications to the Anglican compiler.  The special forms \inline{sample} and \inline{observe} associate values with random variables drawn from a distribution

\begin{bodycode}{}{}
(sample <dist>)
(observe <dist> <value>)
\end{bodycode}

The language additionally provides a data type for sequences of random variables that are not i.i.d., which we refer to as a random process. A random process implements two operations

\begin{bodycode}{}{}
(produce <process>) 
(absorb <process> <value>)
\end{bodycode}

The {\tt produce} primitive returns a distribution object for the next random variable in the sequence. The {\tt absorb} primitive returns an updated random process instance, in which a value has been associated with the next random variable in the sequence. A random process is most commonly used to represent an exchangeably distributed sequence of random variables, but it may be used to represent any sequence of random variables for which it is possible to construct a distribution on the next variable given preceding variables. Random process constructors are customarily identified with uppercase names (e.g.~{\tt CRP}) and, in the same manner as before, are user programmable. 

Finally, the {\tt predict} form may be used at any point in the program to generate labeled output values
\begin{bodycode}{}{}
(predict <label> <expr>)
\end{bodycode}

Given a previously defined query, inference may be performed using the \inline{doquery} macro

\begin{bodycode}{}{}
(doquery algorithm <symbol>
  [<arg1> <arg2> ...] 
  <opt1> <opt2> ...)
\end{bodycode}

The inference algorithm may be specified using an {\tt algorithm} keyword, and any options to the inference algorithm can be supplied as arguments as well. The \inline{doquery} macro constructs a lazy sequence of program execution states that contain predicted values and optionally an importance weight, which may then be consumed and analyzed by, for instance, an outer Clojure or Java program.

\section{Inference}
% !TEX root =  main.tex
%\vspace{-.5cm}
An execution trace  is the sequence of memory states resulting from the sequence of function applications performed during the interpretation of a program.  In probabilistic programming systems like Anglican any variable may be declared as being the output of a random procedure.  Such variables can take different values in independent interpretations of the program.  This leads to a ``many-worlds'' computational trace tree in which, at interpretation time, there is  a branch at every random procedure application.

To define the probability of a single execution trace, first fix an ordering of the exchangeable lines of the program and index the \inline{observe} lines by $n.$  Let $ p(y_n | \theta_{t_n}, \mathbf{x}_{n})$ be the likelihood of the \inline{observe}'d output $y_n$ where $t_n$ is a random procedure type (i.e.~\inline{gamma}, \inline{poisson}, etc.), $\theta_{t_n}$ is its argument (possibly multi-dimensional), and $\mathbf{x}_{n}$ is the set  of all random procedure application results computed before the likelihood of observation $y_n$ is evaluated.  Both the type $t_n$ and the parameter $\theta_{t_n}$ can be functions of any in-scope subset of $\mathbf{x}_n.$   We can then define the probability of an execution trace to be
\begin{eqnarray}
{\tilde p}(\mathbf{y},\mathbf{x}) &\equiv& \prod_{n=1}^N p(y_n | \theta_{t_n}, \mathbf{x}_{n}) {\tilde p}(\mathbf{x}_{n}|\mathbf{x}_{n-1}) \label{eqn:newtracescore}
\end{eqnarray}
 where $\mathbf{y}$ is the set of all \inline{observe}'d quantities, $\mathbf{x}$ is the set of all random procedure application results, and $\sim$ marks distributions which we can only sample.  
%where, again, $\mathbf{x}_{n}$ is the set of all in-scope results of random procedure applications made before the likelihood of observation $y_n$ is to be computed (including both globally- and and locally-scoped random variables, the number and type of which might vary).  

% (the number and type of which may vary from one trace to the next)

The number and type of all random procedure applications performed before the $n$th \inline{observe} may vary in one program trace to the next. We define the probability of the sequence of their outputs $\mathbf{x}_n$ to be
\begin{eqnarray}
\lefteqn{{\tilde p}(\mathbf{x}_{n}|\mathbf{x}_{n-1}) } \label{eqn:pxx}\\
&=& \prod_{k=1}^{|\mathbf{x}_{n}\setminus\mathbf{x}_{n-1}|}p(x_{n,k}|\theta_{t_{n,k}},x_{n,1:(k-1)},\mathbf{x}_{n-1}).\nonumber
\end{eqnarray}
Here $\mathbf{x}_{0} = \emptyset$ is the empty set and $x_{n,j:k}$ are the $j$ to $k$th values generated by random procedure applications in the trace up to observation likelihood computation $n$.  The cardinality of the set $\mathbf{x}_{n}\setminus\mathbf{x}_{n-1}$, notated $|\cdot|$, arises implicitly as the total number of random procedure applications  in a given execution trace.  As before, $\theta_{t_{n,k}}$ are the arguments to, and $t_{n,k}$ the type of, the $(n,k)$th random procedure -- both of which may be functions of subsets of in-scope subsets of variables $x_{n:1,(k-1)} \cup \mathbf{x}_{n-1}$. Note that $\mathbf{x}_{n-1} \subseteq \mathbf{x}_{n}.$  Also note that variable referencing defines a directed conditional dependency structure for the probability model encoded by the program, i.e. $x_{n,k}$ need not (and often cannot due to variable scoping) depend on the outputs of all previous random procedure applications.  %Also note that conditioning on data induces non-trivial statistical dependencies.

We use sampling to explore and characterize the distribution $\tilde{p}(\mathbf{x}|\mathbf{y}) \propto \tilde{p}(\mathbf{y},\mathbf{x})$, i.e.~the distribution of all random procedure outputs that lead to different program execution traces, conditioned on observed data. Related approaches include rejection sampling \cite{goodman2012church}, single-site MH \cite{goodman2012church,wingate2011lightweight}.

All members of the set of all directed probabilistic models with fixed-structure joint distributions can be expressed as probabilistic programs that ``unroll'' in all possible execution traces into an equivalent joint distribution.  As Church-like probabilistic programming frameworks, Anglican included, support recursive procedures and branching on the values returned by random procedures, the corresponding set of models is a superset of the set of all directed graphical models. Other related efforts eschew Turing-completeness and operate on a restricted set of models \cite{hoffman13,pfeffer2001ibal,minkainfer,spiegelhalter1996bugs} where inference techniques other than sampling can more readily be employed.      

%All directed probabilistic graphical models over finite sets of variables can be written as probabilistic programs; in fact, if a probabilistic program has the same set of random choices in every possible execution trace, it can be viewed as a directed graphical model. Church-like probabilistic programming languages, Anglican included, support recursive procedures and branching on the value returned by a random procedure; as a result, execution traces do not correspond directly to directed graphical models. General-purpose inference engines for these expressive languages seem to necessarily depend on sampling.  Other probabilistic programming languages restrict the class of models they can represent, giving up Turing-completeness, and rely on more specialized inference mechanisms \cite{hoffman13,pfeffer2001ibal}, in some cases including compilation into graphical models\cite{minkainfer,spiegelhalter1996bugs}."

% Again, this also need not be true.%Eq.~\ref{eqn:newtracescore}, however, embodies a more expressive model space.

%Let $\mathbf{y}$ denote the set of all observations.

%Next we propose our new method for probabilistic programming inference then review (and clarify) competitive and complimentary prior art.

\subsection{A New Approach}

Towards our new approach to probabilistic programming inference, first consider a standard sequential Monte Carlo (SMC) recursion for sampling from a sequence of intermediate distributions that terminates in $\tilde{p}(\mathbf{x}|\mathbf{y}) \propto \tilde{p}(\mathbf{y},\mathbf{x})$ where $\mathbf{x}$ and $\mathbf{y}$ are as before, and the joint is given by Eq.'s~\ref{eqn:newtracescore}~and~\ref{eqn:pxx}.   Note that a sequence of intermediate approximating distributions can be constructed from any syntactically allowed reordering of 
%\begin{eqnarray*}
%\lefteqn{\tilde{p}(\mathbf{y},\mathbf{x})} \\
%&=& 
\[\tilde{p}(\mathbf{x}_1)\tilde{p}(\mathbf{x}_2|\mathbf{x}_1) \cdots \tilde{p}(\mathbf{x}_{n}|\mathbf{x}_{n-1}) p(y_1|\mathbf{x}_1) \cdots p(y_n|\mathbf{x}_n).\]
%\end{eqnarray*}
%A syntactically valid reordering is any one in which all variables on which term depends are generated before that term is evaluated.
%Throughout the experiments presented in this paper we take the ordering given by the programmer and do not attempt to optimise over allowable expression permutations, although one could reasonably expected to derive improved inference performance from pushing observation likelihoods as far left in this sequence of approximating distributions as possible.  
%
Assume that observation likelihoods are pushed as far left in this sequence of approximating distributions as possible;  however it is clear how to proceed if this is not the case.  Assume we have $1 \leq \ell \leq L$ unweighted samples $\mathbf{x}_{n-1}^{(\ell)} \sim \tilde{p}(\mathbf{x}_{n-1} | y_{1:(n-1)})$
and that from these we will produce approximate samples from  $\tilde{p}(\mathbf{x}_{n} | y_{1:n}).$  To do so via importance sampling we may choose any proposal distribution $q(\mathbf{x}_{n} |\mathbf{x}_{n-1}, y_{1:n})$.  Sampling from this and weighting by the discrepancy between it and the distribution of interest, $\tilde{p}(\mathbf{x}_{n},y_{1:n})$, we arrive at  samples with unnormalized weights 
$\tilde{w}^{(\ell)}_n = \tilde{p}(\hat{\mathbf{x}}_{n}^{(\ell)}, y_{1:n})/q(\hat{\mathbf{x}}_{n}^{(\ell)} |{\mathbf{x}}_{n-1}^{(\ell)}, y_{1:n})$.  
%
%
%
%\begin{eqnarray}
%\tilde{w}^{(\ell)}_n = \frac{\tilde{p}(\hat{\mathbf{x}}_{n}^{(\ell)}, y_{1:n})}{q(\hat{\mathbf{x}}_{n}^{(\ell)} |\mathbf{x}_{n-1}^{(\ell)}, y_{1:n})}. \label{eqn:isweight}
%\end{eqnarray}
%
Here hats notate the difference between weighted and unweighted samples, those with being weighted and vice versa.

This expression simplifies substantially in the ``propose from the prior'' case where the proposal distribution is defined to be the continued interpretation of the program from observation likelihood evaluation $n-1$ to $n$, i.e.~$q(\hat{\mathbf{x}}_{n}^{(\ell)} |\mathbf{x}_{n-1}^{(\ell)}, y_{1:n}) \equiv p(\hat{\mathbf{x}}_{n}^{(\ell)} |\mathbf{x}_{n-1}^{(\ell)})$.  In this case the weight  simplifies to $\tilde{w}^{(\ell)}_n = p(y_n | \hat{\mathbf{x}}_n^{(\ell)}).$  Sampling an unweighted particle set $\mathbf{x}_n^{(\ell)} \sim \sum_\ell w^{(\ell)}_n \delta_{\hat{\mathbf{x}}_{n}^{(\ell)}}$, where $w^{(\ell)}_n = \tilde{w}^{(\ell)}_n/\sum_j \tilde{w}^{(j)}_n$, completely describes SMC for probabilistic program inference.  

The  SMC procedure described is, to first approximation, the inner loop of PMCMC.  It corresponds to a procedure whereby the probabilistic program is interpreted in parallel (possibly each particle in its own thread or process) between observation likelihood calculations.  Unfortunately, SMC with a  finite set of particles is not itself directly viable for probabilistic programming inference for all the familiar reasons: particle degeneracy, inefficiency in models with global, continuous parameters, etc.  

PMCMC, on the other hand, is directly viable.  PMCMC for probabilistic programming inference is a MH algorithm for exploring the space of execution traces that uses SMC proposals internally.  This, unlike prior art, allows sampling of execution traces with changes to potentially many more than one variable at a time.  The particular variant of PMCMC we discuss in this paper is Particle-Gibbs (PG) although we have developed  engines based on other PMCMC variants including particle independent Metropolis Hastings and conditional sequential Monte Carlo too.  PG  works by iteratively re-running SMC, with, on all but the first sweep, reinsertion of a  ``retained'' particle trace into the set of particles at every stage of SMC.  PG is theoretically justified as an MH transition operator that, like the Gibbs operator, always accepts  \cite{andrieu2010particle,holenstein2009particle}. In this paper we describe PMCMC for probabilistic programming inference algorithmically in Alg.\ref{alg:pmcmc} and experimentally demonstrate its relative efficacy for probabilistic programming inference.

In Alg.~\ref{alg:pmcmc} the function \inline{r}$(N,\mathcal{S})$ stands for multi-nomial sample $N$ items from the set of pairs $\mathcal{S} = \{\{w_1,\theta_1\}, \ldots, \{w_M,\theta_M\}\}$, where each element of $\mathcal{S}$ consists of an unnormalized weight $w_m$ and interpreter memory states $\theta_m$.  For each sample value $\theta_m$ returned, the function \inline{r} also returns the original, corresponding unnormalized weight $w_m$.  This kind of weight bookkeeping retains, for all particles, the results of the outermost \inline{observe} likelihood function applications so that the unnormalized weights are available in the retained particle $\{w^*, \mathbf{x}^*\}$ in the next sweep.  %The data structures implicitly by Alg.~\ref{alg:pmcmc}

\begin{algorithm}[H]
\caption{PMCMC for Prob. Prog. Inference}
\label{alg:pmcmc}
\begin{algorithmic}
\STATE $L\gets$ number of particles
\STATE $S\gets$ number of sweeps 
\STATE $\{\tilde{w}_N^{(\ell)},\mathbf{x}_N^{(\ell)}\}\gets$ Run SMC
\FOR{$s < S$}
\STATE $\{\cdot,\mathbf{x}_N^*\}\gets$ \inline{r}$(1,\{1/L,\mathbf{x}_N^{(\ell)}\})$
\STATE $\{\cdot,\mathbf{x}_0^{(\ell)}\}\gets$ initialize $L-1$ interpreters
%\STATE $\tilde{w}_N^* \gets \tilde{w}_N^{(j)}$ where $\mathbf{x}_N^{(j)} = \mathbf{x}_N^*$
\FOR{$d$ $\in$ ordered lines of program}
\FOR{$\ell < L-1$}
	\STATE $\bar{\mathbf{x}}_{n-1}^{(\ell)}\gets$ \inline{fork}$(\mathbf{x}_{n-1}^{(\ell)})$ 
\ENDFOR
\IF {\inline{directive}(d) $==$ ``\inline{assume}''}
	\FOR{$\ell < L-1$}
		\STATE $\bar{\mathbf{x}}_n^{(\ell)}\gets$ \inline{interpret}$(d,\bar{\mathbf{x}}_{n-1}^{(\ell)})$ 
	\ENDFOR
	\STATE $\{\mathbf{x}_n^{(\ell)}\}\gets  \{\bar{\mathbf{x}}_n^{(\ell)}\} \cup \mathbf{x}_n^*$
\ELSIF  {\inline{directive}(d) $==$ ``\inline{predict}''}
	\FOR{$\ell < L-1$}
		\STATE \inline{interpret}$(d,\bar{\mathbf{x}}_{n-1}^{(\ell)})$ 
	\ENDFOR
	 \STATE \inline{interpret}$(d,{\mathbf{x}}_{n-1}^*)$ 
\ELSIF  {\inline{directive}(d) $==$ ``\inline{observe}''}
	\FOR{$\ell < L-1$}
		\STATE $\{\bar{w}_n^{(\ell)},\bar{\mathbf{x}}_n^{(\ell)}\}\gets$ \inline{interpret}$(d,\bar{\mathbf{x}}_{n-1}^{(\ell)})$ 
	\ENDFOR
	\STATE $\mathcal{T}\gets$  \inline{r}$(L-1, \{ \bar{w}_n^{(\ell)}, \bar{\mathbf{x}}_n^{(\ell)}\} \cup  \{ \tilde{w}_n^*, \mathbf{x}_n^*\})$
	\STATE $\{\tilde{w}_n^{(\ell)},\mathbf{x}_n^{(\ell)}\}\gets  \mathcal{T} \cup  \{ \tilde{w}_n^*, \mathbf{x}_n^*\} $
%\STATE $\{\hat{w}_n^{(\ell)},\mathbf{x}_n^{(\ell)}\}\gets$  \inline{r}$(L-1, \{ \bar{w}_n^{(\ell)}, \bar{\mathbf{x}}_n^{(\ell)}\} \cup  \{ \tilde{w}_n^*, \mathbf{x}_n^*\})$
%\vspace{-.3cm}
%\STATE $\{\tilde{w}_n^{(\ell)},\mathbf{x}_n^{(\ell)}\}\gets  \{ \hat{w}_n^{(\ell)}, {\mathbf{x}}_n^{(\ell)}\} \cup  \{ \tilde{w}_n^*, \mathbf{x}_n^*\} $
\ENDIF
\ENDFOR
\ENDFOR
\end{algorithmic}
\end{algorithm}

In  Alg.~\ref{alg:pmcmc} ``Run SMC'' means running one sweep of the $s$ loop with $L$ particles and no retained particle, $d$ is a program line, and \inline{fork}$(\cdot)$ means to copy the entire interpreter memory datastructure (efficient implementations have  characteristics similar to POSIX \inline{fork()} \cite{posixfork}).  The command \inline{interpret}$(d,\cdot)$ means execute line $d$ in the given interpreter.  Only when interpreting an \inline{observe} must the interpreter return a weight, that being result of the outermost apply of the \inline{observe} statement.  Bars indicate temporary data structures, not averages.  All sets are ordered with unions implemented by append operations.

Note that there are more efficient PMCMC algorithms for probabilistic programming inference.  In particular, there is no reason to \inline{fork} unless an \inline{observe} has just been interpreted.  Alg.~\ref{alg:pmcmc} is presented in this form  for expositional purposes.

%In words, PMCMC can be described as follows.  Run SMC as described.  Pick a particle from the final particle set.  For as many iterations as desired (or forever for guarantees), run $L-1$ interpreters until an observation likelihood calculation.  Evaluate the likelihood calculation in each.  Add the retained interpreter (particle) at the same execution point into this set.  Normalise the weights across all $L$ interpreters (the $L-1$ new interpreters plus the retained interpreter) 

%Note that the inherent bias towards short programs shows up differently here than it does in single-site MH approaches.  Here because we are sampling by directly running the program, longer execution paths are, by definition, less likely to 

%Note that one of the reasons we believe that PG prob.~prof.~inference converges faster than random DB is that observes can impact the evolution of a program trace as soon as they are encountered.

%\begin{eqnarray}
%{\tilde p}(\mathbf{y},\mathbf{x}) &\equiv& \prod_{n=1}^N p(y_i | \theta_{t_i}, \mathbf{x}_{1:i}) p_{t_i}(x_i | \theta_{t_i}, \mathbf{x}_{1:(n-1)} )\nonumber \\
%&\times& \prod_{k=1}^{K_i} p_{t_{i,k}}(z_{i,k} | \theta_{t_{i,1}}, z_{i,1}, \ldots z_{{i,k-1}}) \label{eqn:tracescore}
%\end{eqnarray}

\subsection{Random Database}
\label{sec:randomdb}
% !TEX root =  main.tex

We refer to the MH approach to sampling over the space of all traces proposed in \cite{wingate2011lightweight} as ``random database''  (RDB).  A RDB sampler is a MH sampler where a single variable drawn in the course of a particular interpretation of a probabilistic program is modified via a standard MH proposal, and this modification is accepted by comparing the value of the joint distribution of old and new program traces.  For completeness we review RDB here, noting a subtle correction to the acceptance ratio proposed in the original reference which is proper for a larger family of models.

The RDB sampler employs a data structure that holds all random variables $\mathbf{x}$ associated with an execution trace, along with the parameters and log probability of each draw. Note that interpretation of a program is deterministic conditioned on $\mathbf{x}$. A new proposal trace is initialized by picking a single variable $x_{m,j}$ from the $|\mathbf{x}|$ random draws, and resampling its value using a reversible kernel $\kappa(x'_{m,j} | x_{m,j})$. Starting from this initialization, the program is rerun to generate a new set of variables $\mathbf{x}'$ that correspond to a new valid execution trace. In each instance where the random procedure type remains the same, we reuse the existing value from the set $\mathbf{x}$, rescoring its log probability conditioned on the preceding variables where necessary. When the random procedure type has changed, or a new random variable is encountered, its value is sampled in the usual manner. Finally, we compute the probability $p(\mathbf{y} | \mathbf{x'})$ by rescoring each \inline{observe} as needed, and accept with probability
\begin{equation}
\mathrm{min}\left(1,
    \frac
    {p(\mathbf{y} | \mathbf{x'}) p(\mathbf{x'}) q(\mathbf{x} | \mathbf{x'})}
    {p(\mathbf{y} | \mathbf{x}) p(\mathbf{x}) q(\mathbf{x'} | \mathbf{x})}\right)
    .
\end{equation}
In order to calculate the ratio of the proposal probabilities $q(\mathbf{x'} | \mathbf{x})$ and $q(\mathbf{x} | \mathbf{x'})$, we need to account for the variables that were resampled in the course of constructing the proposal, as well as the fact that the sets $\mathbf{x'}$ and $\mathbf{x}$ may have different cardinalities $|\mathbf{x'}|$ and $|\mathbf{x}|$. We will use the (slightly imprecise) notation $\mathbf{x'} \backslash \mathbf{x}$ to refer to the set of variables that were resampled, and let $\mathbf{x'} \cap \mathbf{x}$ represent the set of variables common to both execution traces. The proposal probability is now given by
\begin{equation}
    q(\mathbf{x'} | \mathbf{x})
    =
    \frac{\kappa(x'_{m,j} | x_{m,j})}
         {|\mathbf{x}|}
    \frac{p(\mathbf{x'} \backslash \mathbf{x} \:|\: \mathbf{x'} \cap \mathbf{x})}
         {p(x'_{m,j} | \mathbf{x'} \cap \mathbf{x})} 
    .
\end{equation}
In our implementation, the initialization $x'_{m,j}$ is simply resampled conditioned on the preceding variables, such that $\kappa(x'_{m,j} | x_{m,j}) = p(x'_{m,j} | \mathbf{x'} \cap \mathbf{x})$. The reverse proposal density can now be expressed in a similar fashion in terms of $\mathbf{x} \backslash \mathbf{x'}$ and $\mathbf{x} \cap \mathbf{x'} = \mathbf{x'} \cap \mathbf{x}$, allowing the full acceptance probability to be written as
\begin{equation}
    \frac
    {p(\mathbf{y} | \mathbf{x'}) 
    \: p(\mathbf{x'}) 
    \: |\mathbf{x}| 
    \: p(\mathbf{x} \backslash \mathbf{x'} \:|\: \mathbf{x} \cap \mathbf{x'})}
    {p(\mathbf{y} | \mathbf{x}) 
    \: p(\mathbf{x}) 
    \: |\mathbf{x'}| 
    \: p(\mathbf{x'} \backslash \mathbf{x} \:|\: \mathbf{x'} \cap \mathbf{x})}
    .
\end{equation}

\section{Testing}
\label{sec:tests}
 % !TEX root =  main.tex

%Our experiments check that the mathematical theory driving our algorithmic choices is sound and, second, that the specific implementation of the interpreter correctly implements this theory in an acceptably bug-free way.  

Programming probabilistic program interpreters is a non-trivial software development effort, involving both the correct implementation of an interpreter and the correct implementation of a general purpose sampler.  The methodology we employ to ensure correctness of both involves three levels of testing; 1) unit tests, 2) measure tests, and 3) conditional measure tests.

%These types of testing are a community requirement for probabilistic programming development.  We will contribute to the development of testing standards by publicly releasing our suite of test programs and computed (or analytically derived where possible) ground truth on the web.

%The overall performance of any interpreter depends on choices of memory organisation and the underlying computational system from which features are ``snarfed.''\footnote{\cite{sicp}}  Anglican is reposed on Clojure \cite{clojure}, which is itself reposed on the Java virtual machine \cite{jam}.  This allows a great deal of snarfing\footnote{\cite{steele1983,sicp}}, promotes portability, gives Anglican desirable features such as automatic garbage collection, but is not bare-metal fast.  Matrix operations are performed via Java implementations of BLAS routines \cite{lib-jblas}.  Anglican uses a non-optimised, simple memory organisation.  Our focus was on statistical correctness; reported wall clock times should be viewed as upper bounds. 

\begin{figure*}[!htb]
\begin{center}
\begin{subfigure}[]{.49\textwidth}
	\caption{HMM}
	\includegraphics[width=.49\textwidth]{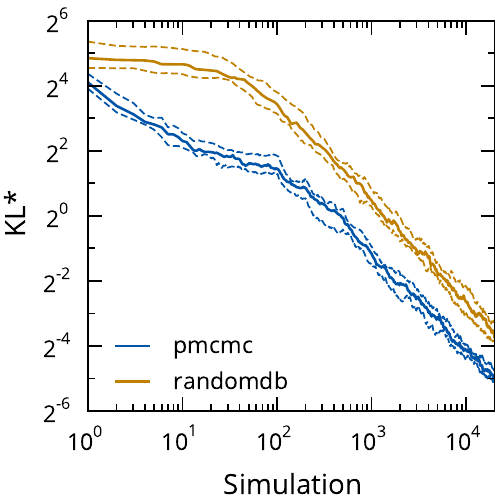}
	\includegraphics[width=.49\textwidth]{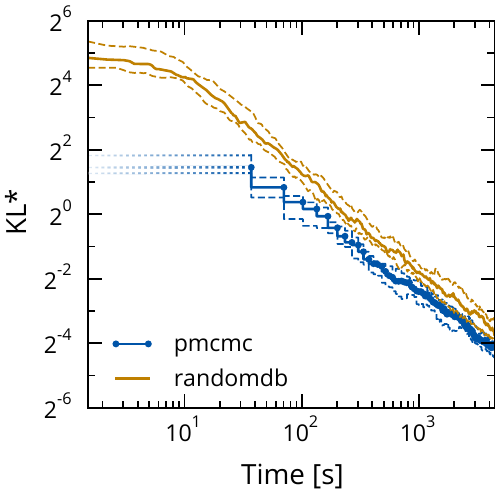}
	\includegraphics[width=.49\textwidth]{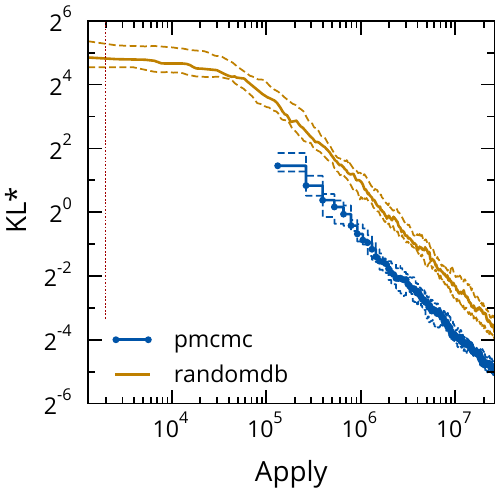}
	\includegraphics[width=.49\textwidth]{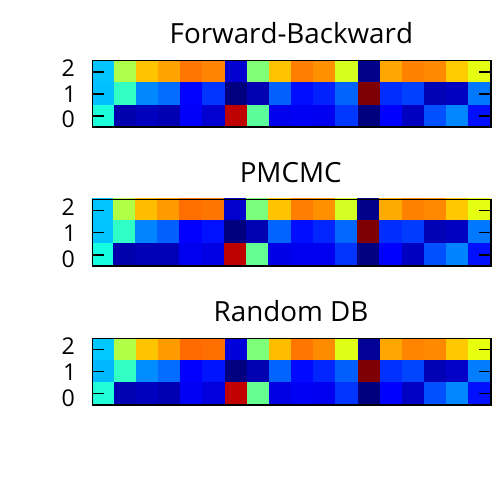}
%	\begin{subfigure}[b]{.49\textwidth}
%	\centering
%	\includegraphics[width=.8\textwidth]{plots/anglican/convergence-comparison/hmm-with-known-obs-and-trans/\commit/gamma-forward-backward.pdf}
%	\includegraphics[width=.8\textwidth]{plots/anglican/convergence-comparison/hmm-with-known-obs-and-trans/\commit/gamma-pmcmc.pdf}
%	\includegraphics[width=.8\textwidth]{plots/anglican/convergence-comparison/hmm-with-known-obs-and-trans/\commit/gamma-randomdb.pdf}
%	\end{subfigure}
	\label{fig:hmm}
\end{subfigure}
\begin{subfigure}[]{.49\textwidth}
	\caption{DP Mixture}
	\includegraphics[width=.49\textwidth]{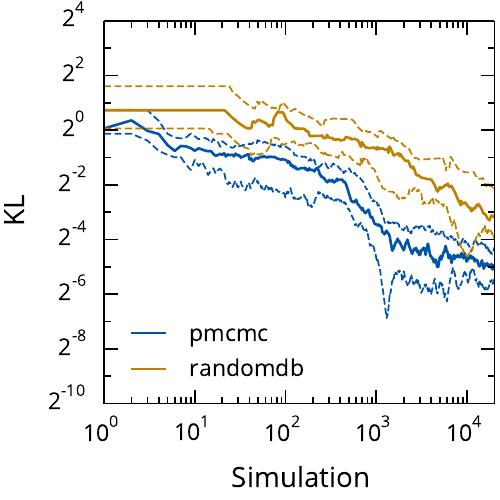}
	\includegraphics[width=.49\textwidth]{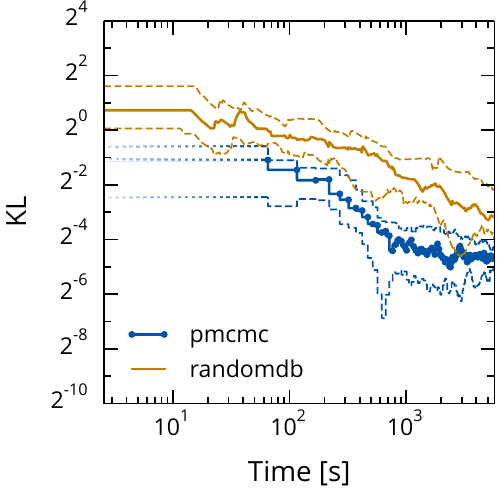}
	\includegraphics[width=.49\textwidth]{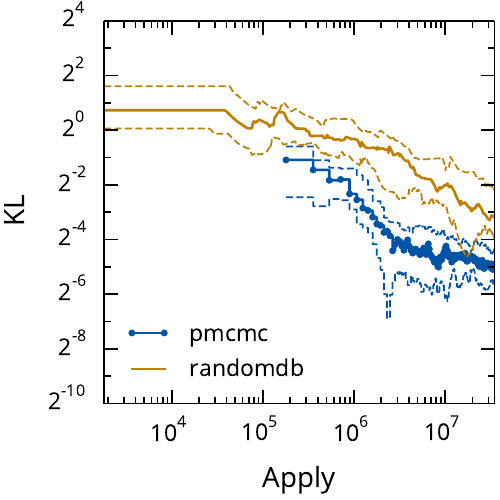}
	\includegraphics[width=.49\textwidth]{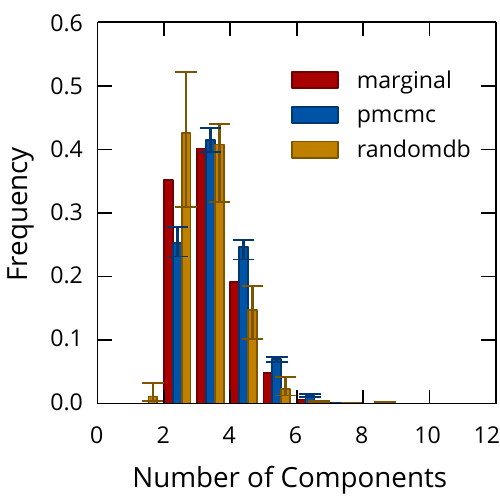}
	\label{fig:crp}
\end{subfigure}
\begin{subfigure}[]{.49\textwidth}
	\caption{Branching}
	\includegraphics[width=.49\textwidth]{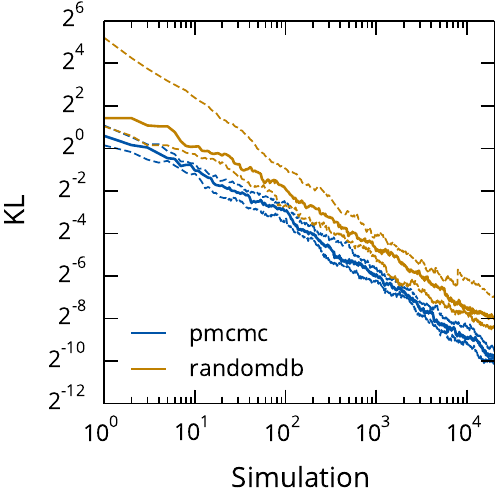}
	\includegraphics[width=.49\textwidth]{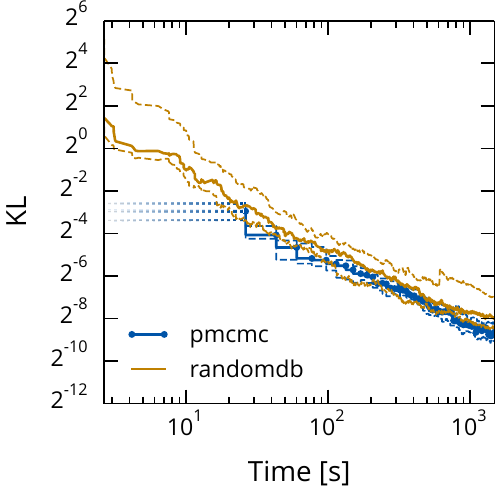}
	\includegraphics[width=.49\textwidth]{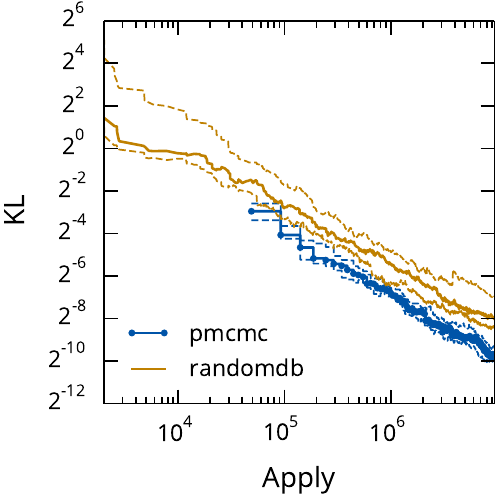}
	\includegraphics[width=.49\textwidth]{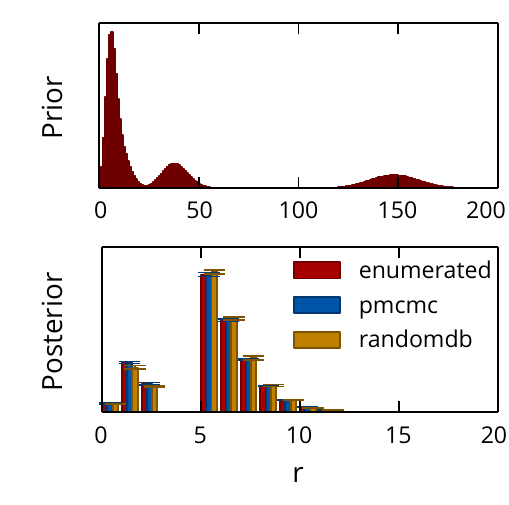}
	\label{fig:branching}
\end{subfigure}
\begin{subfigure}[]{.49\textwidth}
	\caption{Marsaglia}
	\includegraphics[width=.49\textwidth]{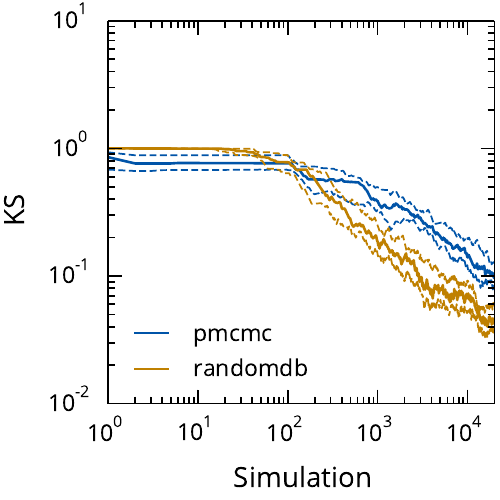}
	\includegraphics[width=.49\textwidth]{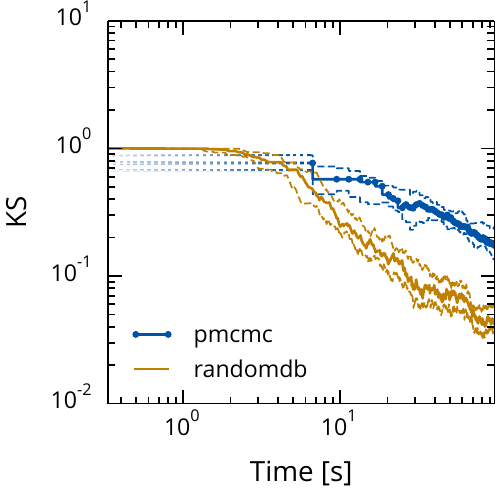}
	\includegraphics[width=.49\textwidth]{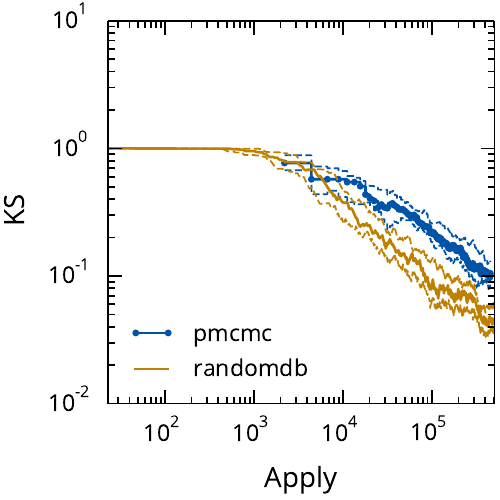}
	\includegraphics[width=.49\textwidth]{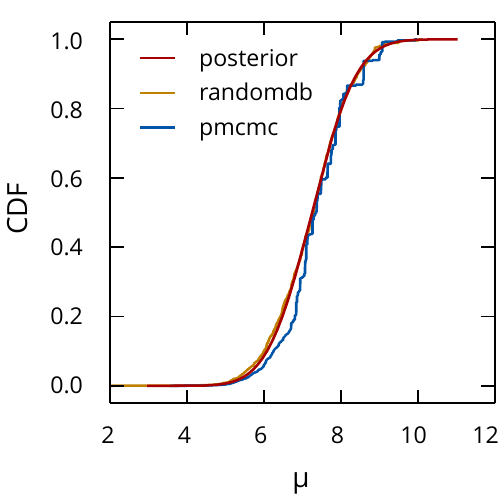}
	\label{fig:marsaglia}
\end{subfigure}
\caption{Comparative conditional measure test performance: PMCMC with 100 particles vs.~RDB.}
\label{fig:convergence}
\end{center}
\end{figure*}

\subsection{Unit and Measure Tests}

In the context of probabilistic programming, unit testing includes verifying that the interpreter correctly interprets a comprehensive set of small deterministic programs.  
%Correct implementations of memoisation (\inline{polya-urn-mem} and its special case \inline{mem}), anonymous functions (\inline{lambda}), and control logic program elements are essential.    Such a suite of test programs was developed to rigorously vet the deterministic components of the interpreter including recursive functions, standard algorithms like sorting, and so forth.  Our interpreter must and did pass all such tests.  
Measure testing involves interpreting short revealing programs consisting of \inline{assume} and \inline{predict} statements (producing a sequence of ancestral, unconditioned samples, i.e.~no \inline{observe}'s).  Interpreter output is tested relative to ground truth, where ground truth is computed via exhaustive enumeration, analytic derivation, or some combination, and always in a different, well-tested independent computational system like Matlab.  Various comparisons of the empirical distribution constructed from the accumulating stream of output \inline{predicts}'s and ground truth are computed; Kulback-Leibler (KL) divergences for discrete sample spaces and Kolmogorov Smirnov (KS) test statistics for continuous sample spaces.  While it is possible to construct distribution equality hypothesis tests for some combinations of test statistic and program we generally are content to accept interpreters for which there is clear evidence of convergence towards zero of all test statistics for all measure tests.  %We call these kinds of tests measure tests to avoid name collision with the unit and distribution testing common to typical software maintenance and release practice.   
Anglican passed all unit and measure tests.  

\subsection{Conditional Measure Tests}

%Initial conformance experiments are unreported due to space; however, these involve writing program that have no \inline{[observe $\ldots$]} statements, rendering interpretation of them essentially ancestral sampling.  Various programs for sampling from known parametric and nonparametric distributions are run and Kolmogorov Smirnov tests \cite{} are employed to test whether the samples generated from interpreting each programs are drawn from the true, known, ground truth distributions.   

%Comparison of different approaches to probabilistic program interpretation can be compared in terms of speed of KL divergence to ground truth relative to computation cost.  As a side note, the need to perform fair comparison of interpreters in terms of computational cost (the latter of the two quantities) suggests that there is a market for a common virtual machine using which all competing probabilistic programming languages can be interpreted and thereby fairly benchmarked.  This is a concern the community is likely to address in due time.

Measure tests involving conditioning provide additional information beyond that provided by measure and unit tests.  Conditioning involves endowing programs with \inline{observe} statements which constrain or weight the set of possible execution traces.     % Ancestral sampling in a probabilistic programming system is comparatively trivial requiring only forward program re-execution; endowing probabilistic program interpreters with 
Interpreting \inline{observe} statements engages the full inference machinery.  Conditional measure test performance is measured in the same way as measure test performance. %;  KL and KS statistics computed between empirical distributions constructed by accumulating  values output by \inline{predict} statements and ground-truth distributions derived by exhaustive enumeration and direct analysis in any combination.
%Classical approaches in the probabilistic programming community involve rejection sampling \cite{} and single-site Metropolis hastings \cite{}.  Our conditional SMC approach is tested both for correctness and in comparison to existing state of the art approaches to single-site Metropolis hastings \cite{}.
  They are also how we compare different probabilistic programming inference engines.

%\subsection{Conformance}
%
%Here too, in all of the following experiments, we compute and show convergence rates for marginals of posterior distributions as a function of computational cost and, where possible or revelatory, as a function of exposed inference parameters of interest.  

\section{Inference Engine Comparison}

We compare PMCMC to RDB measuring convergence rates for an illustrative set of conditional measure test programs.  Results from four such tests are shown in Figure~\ref{fig:convergence} where the same program is interpreted using both inference engines.  %Anglican with both PMCMC and RDB inference engines passes all conditional measure tests but 
PMCMC is found to converge faster for conditional measure test programs that correspond to expressive probabilistic graphical models with rich conditional dependencies. 

The four test programs are: 1) a program that corresponds to state estimation in a hidden Markov model (HMM) with continuous
observations (HMM: Program~\ref{code:hmm}, Figure~\ref{fig:hmm}), 2) a program that corresponds to learning an uncollapsed Dirichlet process (DP) mixture of
Gaussians with fixed hyperparameters (DP Mixture: Program~\ref{code:CRP}, Figure~\ref{fig:crp}),  3) a multimodal branching with deterministic recursion program that cannot be represented as a graphical model in which all possible execution paths can be enumerated (Branching: Program~\ref{code:branching}, Figure~\ref{fig:branching}), and 4) a program that corresponds to inferring the mean of a univariate normal generated via 
an Anglican-coded Marsaglia \cite{marsaglia1964convenient} rejection-sampling algorithm that halts with probability one and generates an unknown number of internal random variables (Marsaglia: Program~\ref{code:marsaglia}, Figure~\ref{fig:marsaglia}).  We refer to 1) and 2) as ``expressive'' models because they have complex conditional dependency structures and 3) and 4) as simple models because the programs encode models with very few free parameters.  1) and 2) illustrate our claims; 3) and 4) are included to document the correctness and completeness of the Anglican implementation while also demonstrating that the gains illustrated in 1) and 2) do not come at too great a cost even for simple programs for which, a priori, PMCMC might be reasonably be expected to underperform.

In Figures~\ref{fig:hmm}-\subref{fig:marsaglia}  there are three panels that report similar style findings across test programs and a fourth that is specific to the individual test program.  In all, PMCMC results are reported for a single-threaded interpreter with 100 particles.  The choice of 100 particles is largely arbitrary; our results are stable for a large range of values. PMCMC is dark blue while RDB is light orange.  We report the 25\% (lower dashed) median (solid) and 75\% (upper dashed) percentiles over 25 runs with differing random number seeds.  We refer to KL/KL$^*$/KS as distances and compute each via a running average of the empirical distribution of \inline{predict} statement outputs to ground truth starting at the first \inline{predict} output.  Note that lower is better.  We define the number of simulations to be the number of times the program is interpreted in its entirety.  For RDB this means that the number of simulations is exactly the number of sampler sweeps; for PMCMC it is the number of particles multiplied by the number of sampler sweeps.  The time horizontal axes report wall clock time; the apply axes report the number of function applications performed by the interpreter.  In the distance vs.~time plots, observed single-threaded PMCMC wall-clock times are reported via filled circles; the left-ward dotted lines illustrate hypothetically what should be achievable via parallelism.  Carefully note that PMCMC requires completing a number of simulations equal to the number of particles (here 100) before emitting batched \inline{predict} outputs.  This means that single-threaded implementations of PMCMC suffer from latency that RDB does not.  Still, for some programs both the quality of PMCMC's \inline{predict} outputs and  PMCMC's  convergence rate is faster even in direct wall clock time comparison to RDB.  
%The apply plots  report one component of computational cost for both inference approaches exactly: interpreter \inline{apply}'s.  The apply plots also illustrate where latency in single-threaded PMCMC arises from: it must do a large amount of computation before each output batch of \inline{predict}'s.  Note that 
PMCMC appears to converge faster for some programs than RDB even relative to the number of function applications.  Equivalent results were obtained relative to \inline{eval} counts.

\subsection{HMM}
{\em New}
\begin{code}{}{code:hmm-cps}
(defquery hmm
  [observations init-dist trans-dists obs-dists]
  (predict
    :states
    (reduce 
      (fn [states obs]
        (let [state (sample (get trans-dists
                                 (peek states)))]
          (observe (get obs-dists state) obs)
          (conj states state)))
      [(sample init-dist)]
      observations)))
\end{code}

{\em Original (deprecated)}
\begin{code}{}{code:hmm}
[assume initial-state-dist (list (/ 1 3) (/ 1 3) (/ 1 3))]
[assume get-state-transition-dist (lambda (s) 
  (cond ((= s 0) (list .1 .5 .4)) ((= s 1) (list .2 .2 .6)) 
        ((= s 2) (list .15 .15 .7))))]
[assume transition (lambda (prev-state) 
  (discrete (get-state-transition-dist prev-state)))]
[assume get-state (mem (lambda (index) 
  (if (<= index 0) (discrete initial-state-dist) 
	           (transition (get-state (- index 1))))))]
[assume get-state-observation-mean (lambda (s) 
	(cond ((= s 0) -1) ((= s 1) 1) ((= s 2) 0)))]
[observe (normal (get-state-obs-mean (get-state 1)) 1) .9]
[observe (normal (get-state-obs-mean (get-state 2)) 1) .8]
$\vdots$
[observe (normal (get-state-obs-mean (get-state 16)) 1) -1]
[predict (get-state 0)]
[predict (get-state 1)]
$\vdots$
[predict (get-state 16)]
\end{code}

The HMM program corresponds to a latent state inference problem in an HMM with three states, one-dimensional Gaussian observations (.9, .8, .7, 0, -.025, 5, 2, 0.1, 0, .13, .45, 6, .2, .3, -1, -1), with known means and variances, transition matrix, and initial state distribution.  The lines of the program were organized with the \inline{observe}'s in ``time'' sequence.

The $KL^*$ axis reports the sum of the Kulback-Leibler divergences between the running sample average state occupancy across all states of a HMM including the initial state and one trailing predictive state 
$KL^*_S = \sum_iD_{KL}( \frac{1}{S}\sum_{s=1}^S \delta_{z_i^{(s)}} || \gamma_i)$.  Here  $\delta_{z_i^{(s)}}(k)$ returns one if simulation $s$ has latent state $z$ at time step $i$ equal to $k$ and $\gamma_i(k)$ is the true marginal probability of the latent state indicator $z_i$ taking value $k$ at time step $i$. % (i.e. $\gamma$ is the usual normalized product of the $\alpha$ and $\beta$ messages from the forward-backward algorithm \cite{forward-backward}).
  The vertical red line in the apply plot indicates the time and number of applies it takes to run forward-backward in the Anglican interpreter.  %In the time plot an additional red dot appears on the first horizontal PMCMC line.  This dot is the wall clock time taken to arrive at the same $KL^*$ by a multi-threaded PMCMC interpreter running with four threads.  PMCMC is theoretically easy to parallelize requiring thread synchronisation only at resampling steps.  Current versions of Anglican include a multi-threaded inference core but work remains to achieve optimal parallelism due, in particular, to memory organisation suboptimality leading to excessive locking overhead.  Even using a single thread, PMCMC outperforms RDB per simulation, wall clock time, and apply count.

The fourth plot shows the learned posterior distribution over the latent state value for all time steps including both the initial state and a trailing predictive time step.  While RDB produces a reasonable approximation to the true posterior, it does so more slowly and with greater residual error.  

\subsection{DP Mixture}

{\em New}
\begin{code}{}{code:CRP-cps}
(defquery crp-mixture
  [observations alpha mu beta a b]
  (let [precision-prior (gamma a b)]
    (loop [observations observations
           state-proc (CRP alpha)
           obs-dists {}
           states []]
      (if (empty? observations)
        (do 
          (predict :states states)
          (predict :num-clusters (count obs-dists)))
        (let [state (sample (produce state-proc))
              obs-dist (get obs-dists
                            state
                            (let [l (sample precision-prior)
                                  s (sqrt (/ (* beta l)))
                                  m (sample (normal mu s))]
                              (normal m (sqrt (/ l)))))]
          (observe obs-dist (first observations))
          (recur (rest observations)
                 (absorb state-proc state)
                 (assoc obs-dists state obs-dist)
                 (conj states state)))))))
\end{code}

{\em Original (deprecated)}
\begin{code}{}{code:CRP}
[assume class-generator (crp 1.72)]
[assume class (mem (lambda (n) (class-generator)))]
[assume var (mem (lambda (c) (* 10 (/ 1 (gamma 1 10)))))]
[assume mean (mem (lambda (c) (normal 0 (var c))))]
[assume u (lambda () (list (class 1) (class 2) $\ldots$   
  (class 9) (class 10)))]
[assume K (lambda () (count (unique (u))))]
[assume means (lambda (i c) 
  (if (= i c) (list (mean c)) 
              (cons (mean i) (means (+ i 1) c) )))]
[assume stds (lambda (i c) 
  (if (= i c) (list (sqrt (* 10 (var c)))) 
    (cons (var i) (stds (+ i 1) c) )))]
[observe (normal (mean (class 1)) (var (class 1))) 1.0]
[observe (normal (mean (class 2)) (var (class 2))) 1.1]
$\vdots$
[observe (normal (mean (class 10)) (var (class 10))) 0]
[predict (u)]
[predict (K)] 
[predict (means 1 (K))]
[predict (stds 1 (K))]
\end{code}

The DP mixture program corresponds to a clustering with unknown mean and variance problem modelled via a Dirichlet process mixture of one-dimensional Gaussians with unknown mean and variance (normal-gamma priors).  The KL divergence reported is between the running sample estimate of the distribution over the number of clusters in the data and the ground truth distribution over the same.  The ground truth distribution over the number of clusters was computed for this model and data by exhaustively enumerating all partitions of the data (1.0, 1.1, 1.2, -10, -15, -20, .01, .1, .05, 0), analytically computing evidence terms by exploiting conjugacy, and conditioning on partition cardinality.  The fourth plot shows the posterior distribution over the number of classes in the data computed by both methods relative to the ground truth.  

This program was written in a way that was intentionally antagonistic to PMCMC. The continuous likelihood parameters were not marginalized out and the \inline{observe} statements were not organized in an optimal ordering.  Despite this, PMCMC outperforms RDB per simulation, wall clock time, and apply count.

%Inference using our new incremental interpreter is demonstrably conformant in that it successfully and correctly estimates the posterior distribution for the following programs: 

\subsection{Branching}

{\em New}
\begin{code}{}{code:branching-cps}
(defn fib [n]
  (loop [a 0 b 1 m 0]
    (if (= m n)
      a
      (recur b (+ a b) (inc m)))))

(with-primitive-procedures [fib]
  (defquery branching []
      (let [count-prior (poisson 4)
            r (sample count-prior)
            l (if (< 4 r)
                6
                (+ (fib (* 3 r))
                   (sample count-prior)))]
        (observe (poisson l) 6)
        (predict :r r))))
\end{code}

{\em Original (deprecated)}
\begin{code}{}{code:branching}
[assume fib (lambda (n) 
  (cond ((= n 0) 1) ((= n 1) 1) 
	      (else (+ (fib (- n 1)) (fib (- n 2))))))]
[assume r (poisson 4)]
[assume l (if (< 4 r) 6 (+ (fib (* 3 r)) (poisson 4)))]
[observe (poisson l) 6]
[predict r]
\end{code}

The branching program has no corresponding graphical model.  It was designed to test for correctness of inference in programs with control logic and execution paths that can vary in the number of sampled values.   It also illustrates mixing in a model where, as shown in the fourth plot, there is a large mismatch between the prior and the posterior, so rejection and importance sampling are likely to be ineffective.  Because there is only one observation and just a single named random variable  PMCMC and RDB should and does achieve essentially indistinguishable performance normalized to simulation, time and apply count.   

%\vspace{1cm}
\subsection{Marsaglia}

{\em New}
\begin{code}{}{code:marsaglia-cps}
(defm marsaglia-normal [mu std]
  (let [u (uniform-continuous -1.0 1.0)]
    (loop [x (sample u)
           y (sample u)]
      (let [s (+ (* x x) (* y y))]
        (if (< s 1.0)
          (+ mu (* std (* x (sqrt (* -2.0 (/ (log s) s))))))
          (recur (sample u) (sample u)))))))

(defquery gaussian-marsaglia
    [observations sigma mu0 sigma0]
    (let [mu (marsaglia-normal mu0 sigma0)
          likelihood (normal mu sigma)]
      (reduce (fn [_ obs]
                (observe likelihood obs))
              nil
              observations)
      (predict :mu mu)))
\end{code}

{\em Original (deprecated)}
\begin{code}{}{code:marsaglia}
[assume marsaglia-normal 
  (lambda (mu std)
    (define x (uniform-continuous -1.0 1.0))
    (define y (uniform-continuous -1.0 1.0))
    (define s (+ (* x x) (* y y))) 
    (if (< s 1) 
      (+ mu (* std (* x (sqrt (* -2.0 (/ (log s) s))))))
      (marsaglia-normal mu std)))]
[assume std (sqrt 2)]
[assume mu (marsaglia-normal 1 (sqrt 5))]
[observe (normal mu std) (+ 8 1)]
[observe (normal mu std) 8]
[predict mu]
\end{code}

Marsaglia is a test program included here for completeness.  It is an example of a type of program for which PMCMC sometimes may not be more efficient.    Marsaglia is the name given to the rejection form of the Box-Muller algorithm \cite{box1958note} for sampling from a Gaussian \cite{marsaglia1964convenient}.  The Marsaglia test program corresponds to an inference problem in which observed quantities are drawn from a Gaussian with unknown mean and this unknown mean is generated by an Anglican implementation of the Marsaglia algorithm for sampling from a Gaussian.  The KS axis is a Kolmogorov-Smirnov test statistic \cite{lilliefors1967kolmogorov} computed by finding the maximum deviation between the accumulating sample and analytically derived ground truth cumulative distribution functions (CDF). Equal-cost PMCMC, RDB, and ground truth CDFs are shown in the fourth plot. 

Because Marsaglia is a recursive rejection sampler it may require many recursive calls to itself.    We conjecture that RDB may be faster than PMCMC here because, while PMCMC pays no statistical cost, it does pay a computational cost for exploring program traces that include many random procedure calls that lead to rejections whereas RDB, due to the implicit geometric prior on program trace length, effectively avoids paying excess computational costs deriving from unnecessarily long traces.  %This conjecture will require substantial effort to formalise and test.  The authors hypothesize that programs of the HMM and DP Mixture example sort are more natural and common but find the Marsaglia example curious.

%All programs end with 
%
%\begin{code}{}{}
%[predict (eval-count)]
%[predict (apply-count)]
%[predict (time)]
%\end{code}

%\begin{figure}[htbp]
%\begin{center}
%\includegraphics[width=.5\textwidth]{fig/muestconvergence.eps}
%\caption{Convergence of Program~\ref{code:gaussian} vs.~number of particles (PMCMC)}
%\label{fig:gau_conv_v_num_particles}
%\end{center}
%\end{figure}

%\begin{figure}[htbp]
%\begin{center}
%\includegraphics[width=.5\textwidth]{fig/branching-prior.eps}
%\caption{Branching program prior (no \inline{[observe \ldots]}'s)}
%\label{fig:branching-prior}
%\end{center}
%\end{figure}

\begin{figure}[htbp]
\begin{center}
\begin{subfigure}[]{.24\textwidth}
	\includegraphics[width=\textwidth]{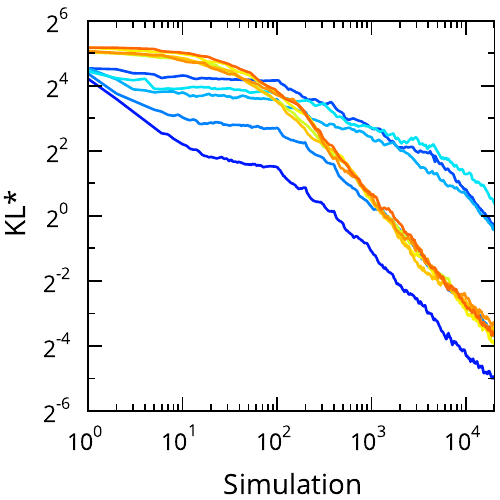}
	\caption{HMM}
	\label{fig:hmm-permutation}
\end{subfigure}\hfill
\begin{subfigure}[]{.24\textwidth}
	\includegraphics[width=\textwidth]{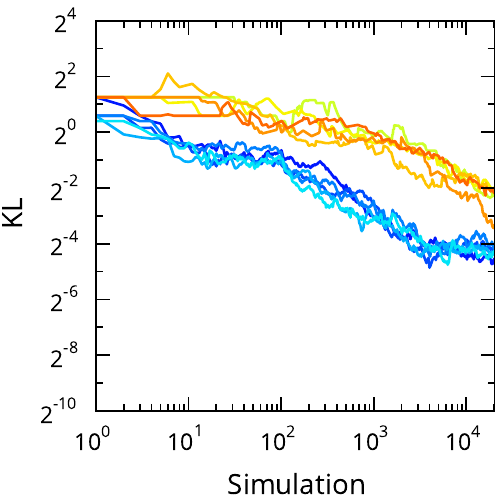}
	\caption{DP Mixture}
	\label{fig:dp-mixture-permutation}
\end{subfigure}
\caption{Effect of program line permutations}
\label{fig:permutation}
\end{center}
\end{figure}

\begin{figure}[htbp]
\begin{center}
\begin{subfigure}[]{.24\textwidth}
	\includegraphics[width=\textwidth]{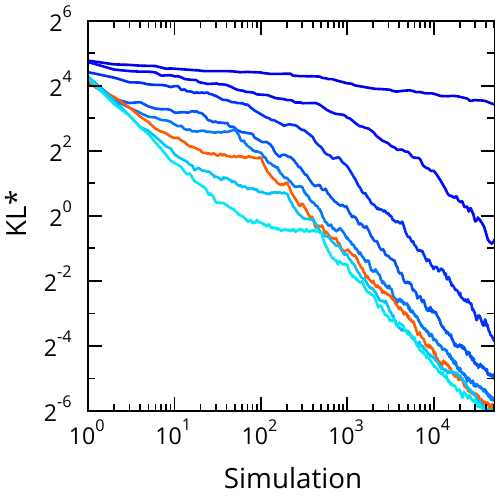}
	\caption{HMM}
	\label{fig:hmm-num-particles}
\end{subfigure}\hfill
\begin{subfigure}[]{.24\textwidth}
	\includegraphics[width=\textwidth]{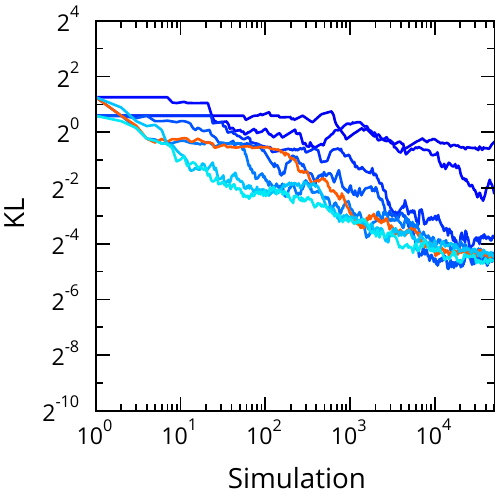}
	\caption{DP Mixture}
	\label{fig:dp-num-particles}
\end{subfigure}
\caption{Effect of particle count on performance}
\label{fig:sweep-num-particles}
\end{center}
\end{figure}

\subsection{Line Permutation}

Syntactically and semantically \inline{observe} and \inline{predict}'s are mutually exchangeable (so too are \inline{assume}'s up to syntactic constraints).  
Given this and the nature of PMCMC it is reasonable to expect that line permutations could effect the efficiency of inference.  We explored this by randomly permuting the lines of the HMM and DP Mixture programs.  The results are shown in Fig.~\ref{fig:permutation} where blue lines correspond to median (out of twenty five) runs of PMCMC for each of twenty five program line permutations (including unmodified (dark) and reversed (light)) and orange are the same for RDB.  For the HMM we found that the natural, time-sequence ordering of the lines of the program resulted in the best performance for PMCMC relative to RDB.  This is because in this ordering the \inline{observe}'s cause re-weighting to happen as soon as possible in each SMC phase of PMCMC.  The effect of permuting code lines interpolates inference performance between optimal where PMCMC is best and adversarial orderings where RDB is instead.  RDB performance is demonstrated here to be independent of the program line ordering.  

The DP Mixture results show PMCMC outperforming RDB on all program reorderings.  Further, it can be seen that the original program ordering was not optimal with respect to PMCMC inference.

While PMCMC presents the opportunity for significant gains in inference efficiency, it does not prevent programmers from seeking to further optimize  performance manually. Programmers can influence inference performance  by reordering program lines, in particular pushing  \inline{observe} statements as near to the front of the program as syntactically allowed, or restructuring programs to lazily rather than eagerly generate latent variables. Efficiency gains via automatic transformations or online adaptation of the ordering may be possible.

\subsection{Number of Particles}

Fig.~\ref{fig:sweep-num-particles} shows the number of particles in the PMCMC inference engine affects performance. Performance improves as a function of the number of particles.  In this plot the red line indicates 100 particles.  Increasing from dark to light the number of particles plotted is 2, 5, 10, 20, 50, (red, 100), 200, and 500. 

%As shown in Fig.~\ref{convergence} it can be seen that the incremental approach to program interpretation converges more rapidly than conventional interpretation approaches in some cases and more slowly in others.  

\section{Discussion}
% !TEX root =  main.tex
%In models with dense conditional dependencies, the PMCMC approach to probabilistic program interpretation appears to converge faster to the true conditional distribution than single-site Metropolis-Hastings methods.
The PMCMC approach to probabilistic program interpretation appears to converge faster to the true conditiona distribution over program execution traces for programs that correspond to expressive models with dense conditional dependencies than single-site Metropolis-Hastings methods.  
That PMCMC converges faster in our tests even after normalizing for computational time is, to us, both surprising and striking.  To reiterate, this includes all the computation done in all the particle executions.% meaning that, per unit energy, PMCMC is more efficient for at least some programs.  That PMCMC is not always faster to converge is not surprising, particularly when the number of random choices is small and the number of constraints is also small. % It is also surprising that, even when compared on such problems and properly normalised for computational cost, it still is competitive. 

Current versions of Anglican include a multi-threaded inference core but work remains to achieve optimal parallelism due, in particular, to memory organization sub-optimality leading to excessive locking overhead.  Using just a single thread, PMCMC surprisingly sometimes outperforms RDB per simulation anyway in terms of wall clock time.

Our specific choice of syntactically forcing \inline{observe}'s to be noisy requires language and interpreter level restrictions and checks.  Hard constraint \inline{observe}'s can be supported in Anglican by exposing a Dirac likelihood  to programmers.  Persisting in not doing so should help programmers avoid writing probabilistic programs where finding even a single satisfying execution trace is NP-hard or not-computable, but also could be perceived as requiring a non-intuitive programming style.  %; indeed, Venture falls back on a slight generalization of this restriction for some of its performance guarantees. %We remain undecided.  

As explored in concurrent work on Venture, there may be opportunities to improve inference performance by partitioning program variables, either automatically or via syntax constructs, and treating them differently during inference. Then, for instance, some could be sampled via plain MH, some via conditional SMC.  One simple way to do this would be to combine conditional SMC and RDB MH. 
 
%We hope that our probabilistic program interpretation will significantly expand the usefulness and direct applicability of the probabilistic programming methodology.  
%Anecdotally, more than once we had the experience that Anglican pointed out subtle bugs in the code we used to produce ground truth! 

\subsubsection*{Acknowledgments}

We thank Xerox and Google for their generous support.  We also thank Arnaud Doucet, Brooks Paige, and Yura Perov for helpful discussions about PMCMC and probabilistic programming in general.  

%\subsubsection*{References}

\newpage

\bibliographystyle{plainnat}
\small{
\bibliography{anglican}
}

\newpage

%\section{Supplementary Materials}
%
%\subsection{Models}
%\input{models}
%
%\subsection{Programs}
%\input{programs}
%\newpage
%\subsection{Reserved Keywords}
%\begin{figure*}[htbp]
%\begin{code}{Anglican expression syntax and built-in functions}{}
%(define (function first-argument $\ldots$ last-argument) body)
%(define variable value)
%(quote string)
%(if predicate consequent alternative)
%(lambda (first-argument $\ldots$ last-argument) body)
%(begin first-statement  $\ldots$ last-statement)
%(inc integer)
%(dec integer)
%(not boolean)
%(car list)
%(cdr list)
%(cons item list)
%(null? value)
%(+ number number)
%(* number number)
%(- number number)
%(/ number number)
%(= value value)
%(> number number)
%(>= number number)
%(< number number)
%(<= number number)
%(power number number)
%(mod number number)
%(list first-value $\ldots$ last-value)
%(first list)
%(rest list)
%(or list-of-booleans)
%(and list-of-booleans)
%(cos number)
%(sin number)
%(gamma alpha beta)
%(poisson lambda)
%(normal mu sigma)
%(dirichlet (list first-pseudocount $\ldots$ last-pseudocount))
%(exponential lambda)
%(flip weight)
%(uniform-continuous min-number max-number)
%(discrete-continuous min-integer max-integer)
%(beta alpha beta)
%(discrete (list first-pseudocount $\ldots$ last-pseudocount))
%(categorical (list first-pseudocount $\ldots$ last-pseudocount))
%(natural)
%(polya-urn-mem concentration function)
%(mem function)
%(dirichlet-multinomial (list first-pseudocount $\ldots$ last-pseudocount))
%(beta-bernoulli (list zero-pseudocount one-pseudocount))
%$\vdots$
%\end{code}
%\end{figure*}
%
%
%\section{Related Work}
%\input{relatedwork}

\end{document}